\pdfoutput=1
\documentclass[11pt]{article}

\usepackage{EMNLP2022}

\usepackage{times}
\usepackage{latexsym}
\usepackage{graphicx}
\usepackage{booktabs,tabularx,enumitem,ragged2e}
\usepackage{times}
\usepackage{amsmath}
\usepackage{latexsym}
\usepackage{multirow}
\usepackage{xcolor}
\usepackage{lscape}
\usepackage{subfig}
\usepackage{graphicx}
\usepackage{amsmath,amsthm,amssymb}
\usepackage[boldmath]{numprint}
\usepackage{color, colortbl}

\definecolor{LightGreen}{HTML}{d6f5d6}
\definecolor{LightRed}{HTML}{ffcccc}

\newcommand{\centered}[1]{\begin{tabular}{l} #1 \end{tabular}}

\usepackage[T1]{fontenc}
\usepackage[utf8]{inputenc}

\usepackage{microtype}

\usepackage{inconsolata}

\title{Are Hard Examples also Harder to Explain? A Study with Human and Model-Generated Explanations}

\author{Swarnadeep Saha$^1$ \quad Peter Hase$^1$ \quad Nazneen Rajani$^2$ \quad Mohit Bansal$^1$
\\ 
  $^1$UNC Chapel Hill \quad $^2$Hugging Face\\
\texttt{\{swarna, peter, mbansal\}@cs.unc.edu}, \texttt{nazneen@huggingface.co}
}

\begin{document}
\maketitle
\begin{abstract}

Recent work on explainable NLP has shown that few-shot prompting can enable large pre-trained language models (LLMs) to generate grammatical and factual natural language explanations for data labels. In this work, we study the connection between explainability and sample hardness by investigating the following research question -- ``Are LLMs and humans equally good at explaining data labels for both easy and hard samples?'' We answer this question by first collecting human-written explanations in the form of generalizable commonsense rules on the task of Winograd Schema Challenge (Winogrande dataset). We compare these explanations with those generated by GPT-3 while varying the hardness of the test samples as well as the in-context samples. We observe that (1) GPT-3 explanations are as grammatical as human explanations regardless of the hardness of the test samples, (2) for easy examples, GPT-3 generates highly supportive explanations but human explanations are more generalizable, and (3) for hard examples, human explanations are significantly better than GPT-3 explanations both in terms of label-supportiveness and generalizability judgements. We also find that hardness of the in-context examples impacts the quality of GPT-3 explanations. Finally, we show that the supportiveness and generalizability aspects of human explanations are also impacted by sample hardness, although by a much smaller margin than models.\footnote{Supporting code and data are available at \url{https://github.com/swarnaHub/ExplanationHardness}.}

\end{abstract}

\begin{figure*}[t]
\centering
\resizebox{\textwidth}{!}{%
\begin{tabular}{p{0.45\linewidth} | l | p{0.31\linewidth} | p{0.37\linewidth} }
\toprule
\textbf{Sentence} & \textbf{Options (\underline{Answer})} & \textbf{Human Explanation} &\textbf{GPT-3 Explanation} \\ \midrule
\rowcolor{LightGreen}
Katrina had the financial means to afford a new car while Monica did not, since \_ had a high paying job. & \underline{Katrina}/Monica & If X has a high paying job, then X is able to afford expensive purchases &  If X has a higher paying job than Y, then X would have more money to afford a new car. \\ \midrule
\rowcolor{LightGreen}
Angela tried to calm Carrie's nerves at the airport because \_ was scared of flying in airplanes. & Angela/\underline{Carrie} & If X is afraid and Y is not, then Y can help calm X down. &	 If X is scared of flying, then Y will try to calm X down. \\ \midrule
\rowcolor{LightGreen}
Donald was always reading while Kenneth did not, as \_ liked to read books all the time. &	\underline{Donald}/Kenneth &	If X enjoys reading more than Y, X will read more books than Y	& If X reads more than Y, then X likes to read books all the time. \\ \midrule
\rowcolor{LightRed}
Justin spent less on dog food than Eric because \_ had a puppy dog at their house. & \underline{Justin}/Eric & If X has a puppy and Y has a dog, then a puppy eats less so X spends less on dog food. &	 If X has a Y at their house, then X will spend less on Y. \\ \midrule
\rowcolor{LightRed}
I grabbed a pile of clothes from the basket that contained a T-shirt and a Sweater, since it was hot outside I put on the \_ to wear. &	Sweater/\underline{T-Shirt} & If X is cooler than Y, then X will be chosen when it is hot outside. & If X is hot, then Y should be worn. \\ \midrule
\rowcolor{LightRed}
Helen was curious about how the ball broke the window, because the \_ was hard. &	ball/\underline{window}	&	If it is surprising that X is broken then X must have been hard, otherwise it wouldn't be surprising that Y broke X. &	 If X is hard and Y isn't, then Y is more likely to break when hit by X. \\
\bottomrule      
\end{tabular}%
\vspace{-10pt}
}
\caption{Representative examples of explanations for Winograd Schema written by humans and generated by GPT-3 for easy (first 3 rows) and hard examples (last 3 rows). For easy examples, GPT-3 explanations are almost as good as humans, although less generalizable. For example, humans can generalize `cars' to `expensive purchases' while the model does not. For hard examples, GPT-3 explanations are often much worse than human ones.\label{tab:contrast_easy_hard}
\vspace{-10pt}
}

\end{figure*}

\section{Introduction}

Prior work on explainable NLP~\cite{wiegreffe2021teach} has explored different forms of explanations ranging from extractive rationales~\cite{zaidan2007using, deyoung2020eraser}, semi-structured, and structured explanations~\cite{jansen2018worldtree, mostafazadeh2020glucose, saha2021explagraphs} to free-text explanations~\cite{camburu2018snli}. Due to the flexibility of free-text explanations, they have emerged as a popular form of explanations with multiple benchmarks developed around them, as well as models that generate such explanations using seq2seq language models~\cite{ehsan2018rationalization, camburu2018snli, rajani2019explain, narang2020wt5}. Few-shot prompting~\cite{radford2019language, schick2021exploiting} with Large Language Models (LLMs) like GPT-3~\cite{brown2020language} has been shown to produce highly fluent and factual natural language explanations that are often preferred over crowdsourced explanations in existing datasets~\cite{wiegreffe2021reframing}. However, past work has not yet explored a critical dimension of datapoint-level explanations, which is how \emph{hard} the data point is to classify correctly. Given recent work on measuring \emph{hardness} of individual data points~\cite{swayamdipta2020dataset}, we study how sample hardness influences both LLMs' and humans' ability to explain data labels. In summary, we are interested in investigating the following three research questions: 
\begin{enumerate}
[nosep, wide=0pt, leftmargin=*, after=\strut]
\item \textbf{RQ1.} \textit{Do LLMs explain data labels as well as humans for both easy and hard examples?} 
\item \textbf{RQ2.} \textit{How much do LLM explanations vary based on the size and the hardness of the retrieval pool for choosing in-context samples?}
\item \textbf{RQ3.} \textit{Are humans equally good at explaining easy and hard examples?}
\end{enumerate}

As a case study, we investigate these questions for a classical commonsense reasoning task, Winograd Schema Challenge~\cite{levesque2012winograd} on a large-scale dataset, Winogrande~\cite{sakaguchi2020winogrande} (examples in Fig.~\ref{tab:contrast_easy_hard}). We first collect generalizable rule-based explanations from humans like ``If X is larger than Y, then X does not fit in Y.''. To measure data \emph{hardness}, we use \textit{Data Maps} \cite{swayamdipta2020dataset}, an approach based on the training dynamics of a classification model. Similar to~\citet{wiegreffe2021reframing}, we generate post-hoc explanations by conditioning on the answer leveraging GPT-3 with in-context learning. We perform human evaluation of the crowdsourced and model-generated explanations and compare them on the basis of `grammaticality', `supportiveness' and `generalizability'. In summary, we report the following findings:

\begin{itemize}[nosep, wide=0pt, leftmargin=*, after=\strut]
\item LLM-generated explanations match the grammaticality/fluency of human-written explanations regardless of the hardness of test samples.
\item For easy examples, both models and humans write `supportive' explanations, but humans write more `generalizable' explanations that can explain multiple similar data points. For hard examples, humans write explanations that are not only more `generalizable' but also significantly more `supportive' of the label.
\item While choosing in-context examples, factors like size and hardness of the retrieval pool affect the quality of model-generated explanations.
\item Humans, while much better than models in explaining hard examples, also struggle with writing generalizable explanations for these points, succeeding only about 2/3rd of the time.
\end{itemize}

\section{Method and Experimental Setup}
Our method first estimates hardness of the samples using Data Maps~\cite{swayamdipta2020dataset} and then chooses a subset of easy, medium, and hard examples, for which we collect human-written explanations and generate explanations from a state-of-the-art model. Next, we answer our research questions by comparing the explanations against multiple granular evaluation axes.
\paragraph{Data Maps.} We estimate sample hardness via a model-based approach\footnote{We do not rely on human annotations for hardness quantification because of its subjectivity. Data Maps also provide a hardness ranking of the samples, which might be difficult to obtain from humans.} called Data maps~\cite{swayamdipta2020dataset}. Data Maps characterize points $x_i$ in a dataset along two dimensions according to a classifier's behavior during training: (1) confidence $\hat{\mu}_i$ which measures the mean model probability of the true label $y_i^*$ across $E$ epochs, and (2) variability $\hat{\sigma}_i$ which measures the standard deviation of the model probability of the true label across epochs.
\vspace{-7pt}
\begin{equation}
\small
\left.\begin{aligned}
    \hat{\mu}_i &= \frac{1}{E} \sum_{e=1}^E p_{\theta^{(e)}} (y_i^* | x_i) \nonumber \\
    \hat{\sigma}_i &= \sqrt{\frac{\sum_{e=1}^E (p_{\theta^{(e)}} (y_i^* | x_i) - \hat{\mu}_i)^2}{E}} \nonumber \\
\end{aligned}\right.
\end{equation}
where $p_{\theta^{(e)}}$ denotes the model's probability with parameters $\theta^{(e)}$ at the end of the $e^{th}$ epoch. These two metrics give rise to different portions in the dataset including \textit{easy-to-learn} examples where the model consistently predicts the sample correctly across epochs (high confidence, low variability), \textit{hard-to-learn} examples where the model rarely predicts the sample correctly (low confidence, low variability) and \textit{ambiguous} examples where the model is indecisive about its predictions (high variability). We fine-tune RoBERTa-large~\cite{liu2019roberta} on the Winogrande dataset to compute the confidence and variability of each training sample in the dataset. The two metrics are then used to rank the samples from \textit{easy} to \textit{hard} (most confident to least confident) and \textit{least-ambiguous} to \textit{most-ambiguous} (least variable to most variable). As discussed later, we choose a subset of these examples to compare human and model-generated explanations.
\paragraph{Explanations for Winograd Schema.} Next, we define the structure of explanations for the Winograd Schema Challenge~\cite{levesque2012winograd}. Specifically, these are semi-structured if-then commonsense rules as shown in Fig.~\ref{tab:contrast_easy_hard}. This characterization of explanations allows us to (1) capture generalizable commonsense knowledge via placeholders X (and Y) capable of explaining a number of similar data points, (2) enforce the common structural form of an if-then rule for all data points in this task, while still maintaining the flexibility of free-text explanations (see Fig.~\ref{tab:contrast_easy_hard} for some examples), (3) ensure non-trivial explanations that do not leak the label~\cite{hase2020leakage}, with the aim of avoiding explanations that only repeat the label without providing generalizable background knowledge (a common issue in past explanation datasets), (4) evaluate explanation properties with reduced human subjectivity due to their semi-structural form.

\paragraph{Human Explanation Collection.} Using the above criteria for constructing explanations (see detailed instructions in Fig.~\ref{fig:explanation_collection}), we collect human-written explanations on Amazon Mechanical Turk. In order to ensure that the explanations do not explicitly leak the label, the annotators are asked to write explanations in the form of generalizable commonsense rules consisting of placeholders X (and Y) without mentioning the actual options. We collect explanations for 500 easiest and 500 hardest samples, along with 100 examples with medium hardness (around the median confidence). We do not collect explanations separately for least and most ambiguous samples because ambiguity correlates strongly with hardness, i.e., the least ambiguous examples are often the easiest while the most ambiguous examples are also typically the hardest.

\paragraph{Explanation Generation via GPT-3. } Next, we select GPT-3~\cite{brown2020language} as a representative candidate of today's NLP model landscape to generate explanations from. For each set of 500 easy and hard samples, we randomly split them into 400 samples for retrieving in-context samples and 100 samples for testing. We generate explanations for the test samples using the largest (175B) ``text-davinci-002'' InstructGPT model of GPT-3 by conditioning on the context and the gold label (as shown in Fig.~\ref{tab:prompt_example}). The in-context samples are chosen by computing the embeddings of the test sample and the retrieval samples using Sentence BERT~\cite{reimers2019sentence} and selecting the top-k samples (see Appendix~\ref{sec:top_k_samples} for examples). We set $k$ to 5 in our experiments. Further details of our prompting method are in Appendix \ref{sec:prompt}. 
\paragraph{Explanation Evaluation.} Having obtained human and model explanations, we now describe their evaluation process. Due to the limitations of automatic metrics for evaluating explanation quality~\cite{clinciu2021study}, we follow~\citet{wiegreffe2021reframing} to conduct human evaluation of both crowdsourced and GPT-3 explanations on MTurk based on three attributes -- \textit{grammaticality}, \textit{supportiveness}, and \textit{generalizability}. When evaluating explanations for \textit{grammaticality}, we evaluate their syntax and fluency while ignoring spelling mistakes and typos (which also hardly ever appear in model explanations). Given the semi-structured nature of our explanations, we evaluate \textit{supportiveness} as whether, when appropriately replacing X and Y with the two options, the explanation answers the question ``Why does this point receive the label it does?''~\cite{miller2019explanation}. Lastly, we evaluate \textit{generalizability} as how applicable the explanation is for other samples with different X and Y. We maintain a trained pool of annotators for both explanation authoring and verification while ensuring that they do not verify their own data. Each explanation is evaluated by 3 different annotators and the final results are obtained by majority voting. We report moderate inter-annotator agreement scores of Krippendorff’s $\alpha$~\cite{krippendorff2011computing} between 0.4-0.6, details of which are discussed in Appendix~\ref{sec:crowdsource}.

\begin{figure*}
    \centering
    \begin{minipage}{0.33\textwidth}
        \centering
        \includegraphics[width=\linewidth]{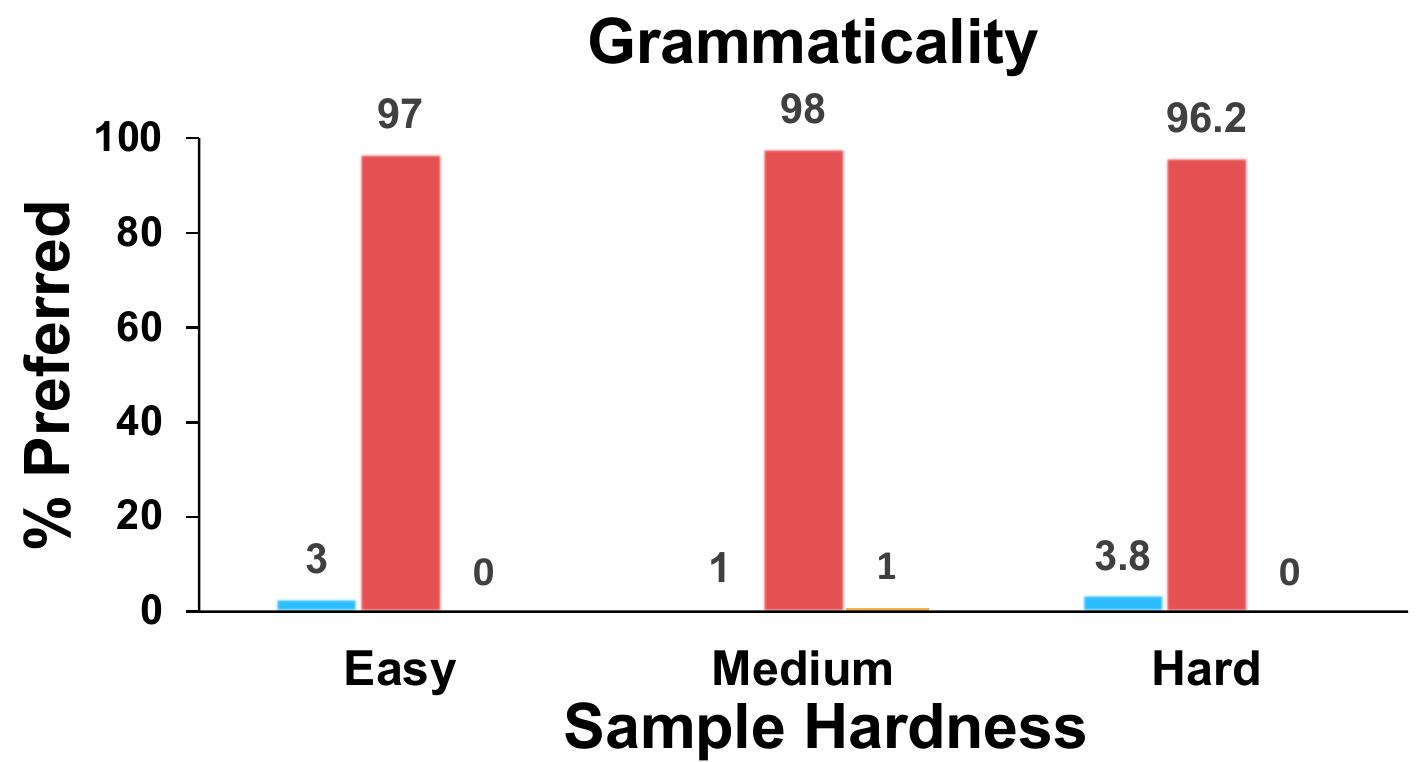} % 
    \end{minipage}\hfill
    \begin{minipage}{0.33\textwidth}
        \centering
        \includegraphics[width=\linewidth]{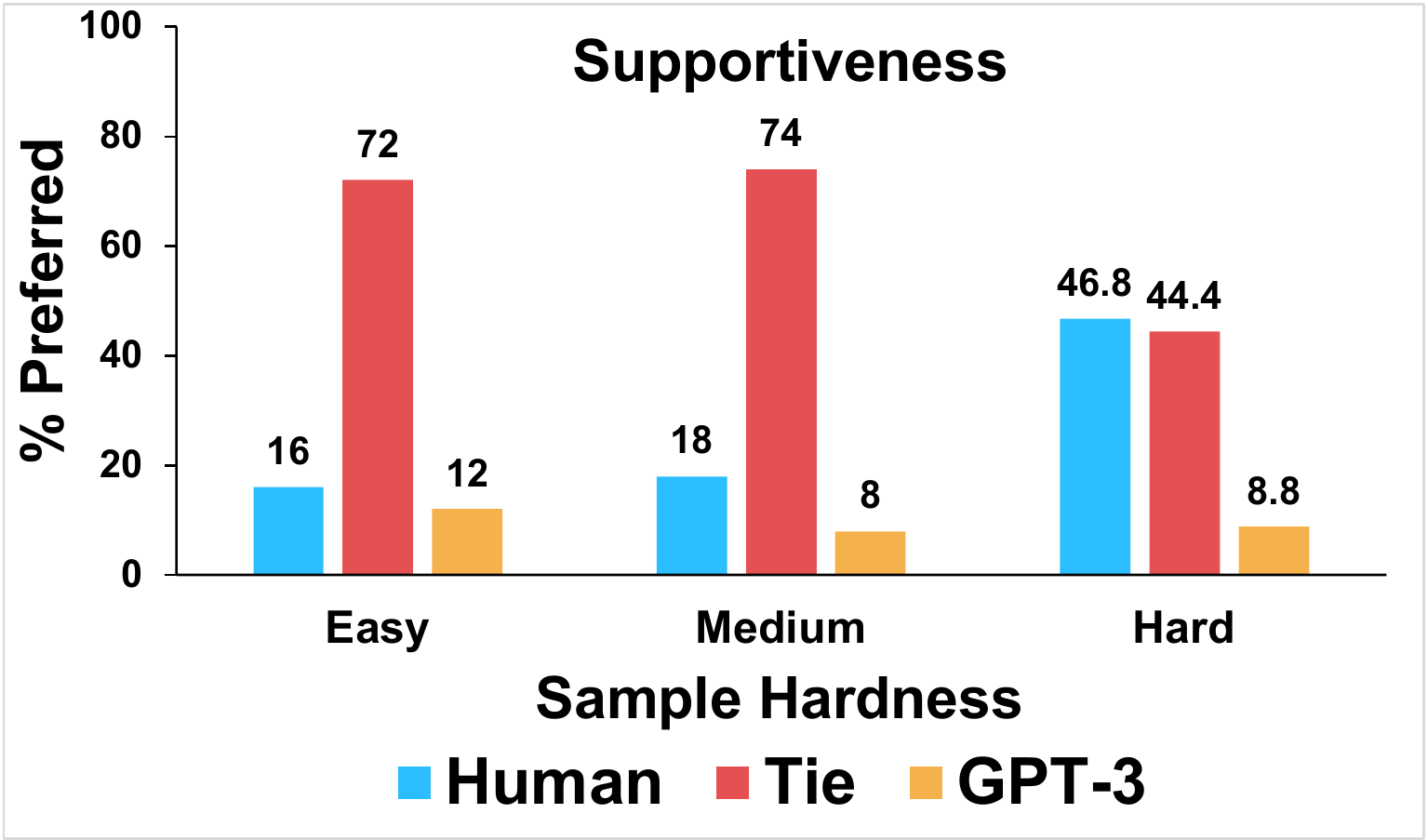} %
    \end{minipage}
    \begin{minipage}{0.33\textwidth}
    \raggedleft
        \includegraphics[width=\linewidth]{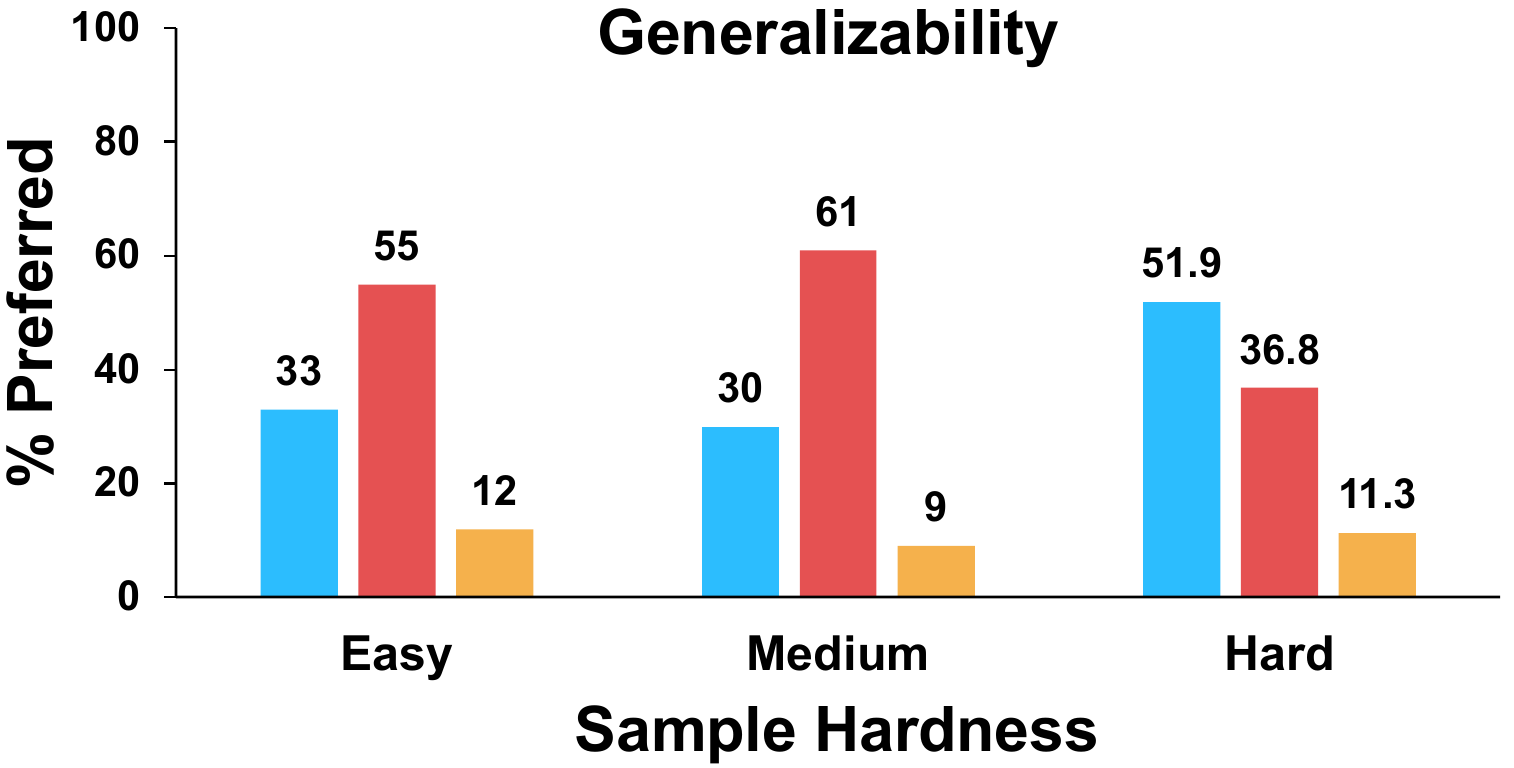} %
    \end{minipage}
    \vspace{-5pt}
    \caption{\label{fig:contrast_easy_hard}Head-to-head comparison of human and GPT-3 explanations for easy, medium and hard examples along the axes of grammaticality, supportiveness and generalizability.}
    \vspace{-10pt}
\end{figure*}

\section{Results}

\subsection{RQ1: Do LLMs explain data labels as well as humans for both easy and hard examples?}

In Fig.~\ref{fig:contrast_easy_hard}, we compare the human and GPT-3 explanations for easy, medium, and hard\footnote{Some hard examples can have incorrect labels~\cite{swayamdipta2020dataset}. When collecting explanations from humans, we ask if they agree with the label (see Fig.~\ref{fig:explanation_collection}). If they do not, we discard such examples (about 14\%) from evaluation.} examples along the three axes. We observe that GPT-3 is not only able to learn the if-then structure of the explanations but also matches humans in terms of generating grammatically fluent explanations, regardless of the sample hardness. For easy examples, GPT-3 explanations are almost as supportive of the label as human explanations, sometimes even outperforming humans. However, humans are typically better at writing more generalizable explanations that apply to broader contexts (see examples 1-3 in Fig.~\ref{tab:contrast_easy_hard}). For hard examples, GPT-3 often fails to generate sufficiently supportive explanations and hence significantly underperforms humans in more than 46\% of the cases (see examples 4-6 in Fig.~\ref{tab:contrast_easy_hard} and Appendix~\ref{sec:analysis-gpt3} for some common errors). This, in turn, also hurts the generalizability aspect of the model-generated explanations. Medium-hard examples show a trend similar to easy examples because their confidence values are much closer to the easy examples than the hard ones. 

\noindent \textbf{Significance Testing.} Pertaining to the above results, we further use a non-parametric bootstrap test~\cite{efron1994introduction} to evaluate whether the human win-rate differs significantly from the model win-rate, while treating ties as neutral. 
We encode human wins as 1, model wins as -1, and ties as 0 and test whether the average score is not equal to 0 (meaning that the win-rate differs between human and model). In summary, for easy and medium samples, humans' generalizability is significantly better than the model’s (difference in win rate is 20 points with $p{<}$0.001), while for hard samples, both humans' generalizability and supportiveness are better than the model’s (differences in win rates are 0.38 and 0.4 respectively, with $p{<}$1e-4). Next, for grammaticality, we test if GPT-3 explanations matches human explanations within a win-rate threshold of $\pm r$ points. For a threshold of $r{=}0.1$ (testing that grammaticality win-rates are within 10 percentage points of each other), we obtain $p{<}$0.05, and a threshold of $r{=}0.15$ yields $p{<}$1e-4. This suggests that the model’s grammaticality significantly matches human’s for \textit{threshold} values around 0.1.

\begin{figure*}
    \centering
    \begin{minipage}{\columnwidth}
        \centering
        \includegraphics[width=\linewidth]{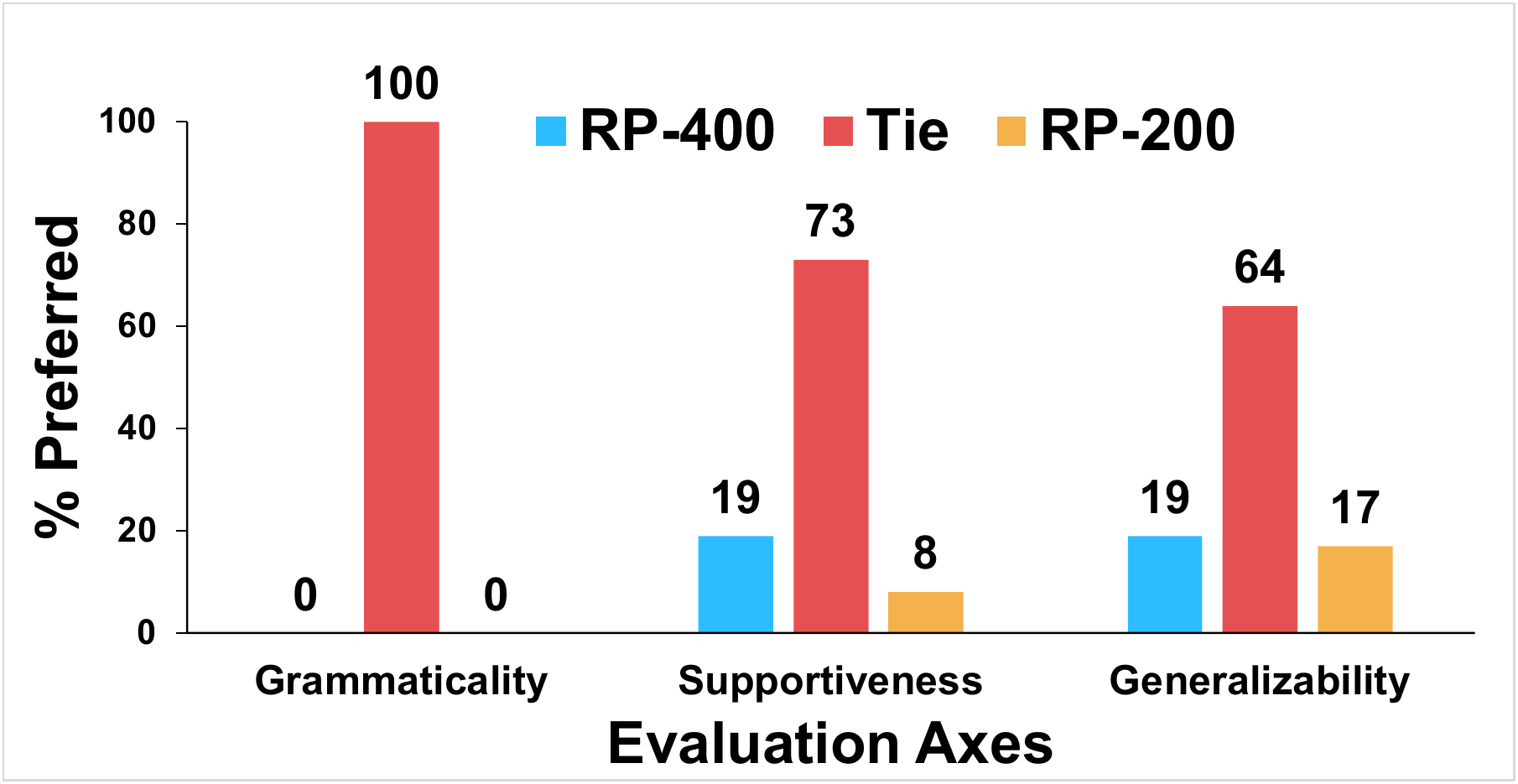}
 \small{(a) Size of Retrieval Pool}        
    \end{minipage}\hfill
    \begin{minipage}{\columnwidth}
        \centering
        \includegraphics[width=\linewidth]{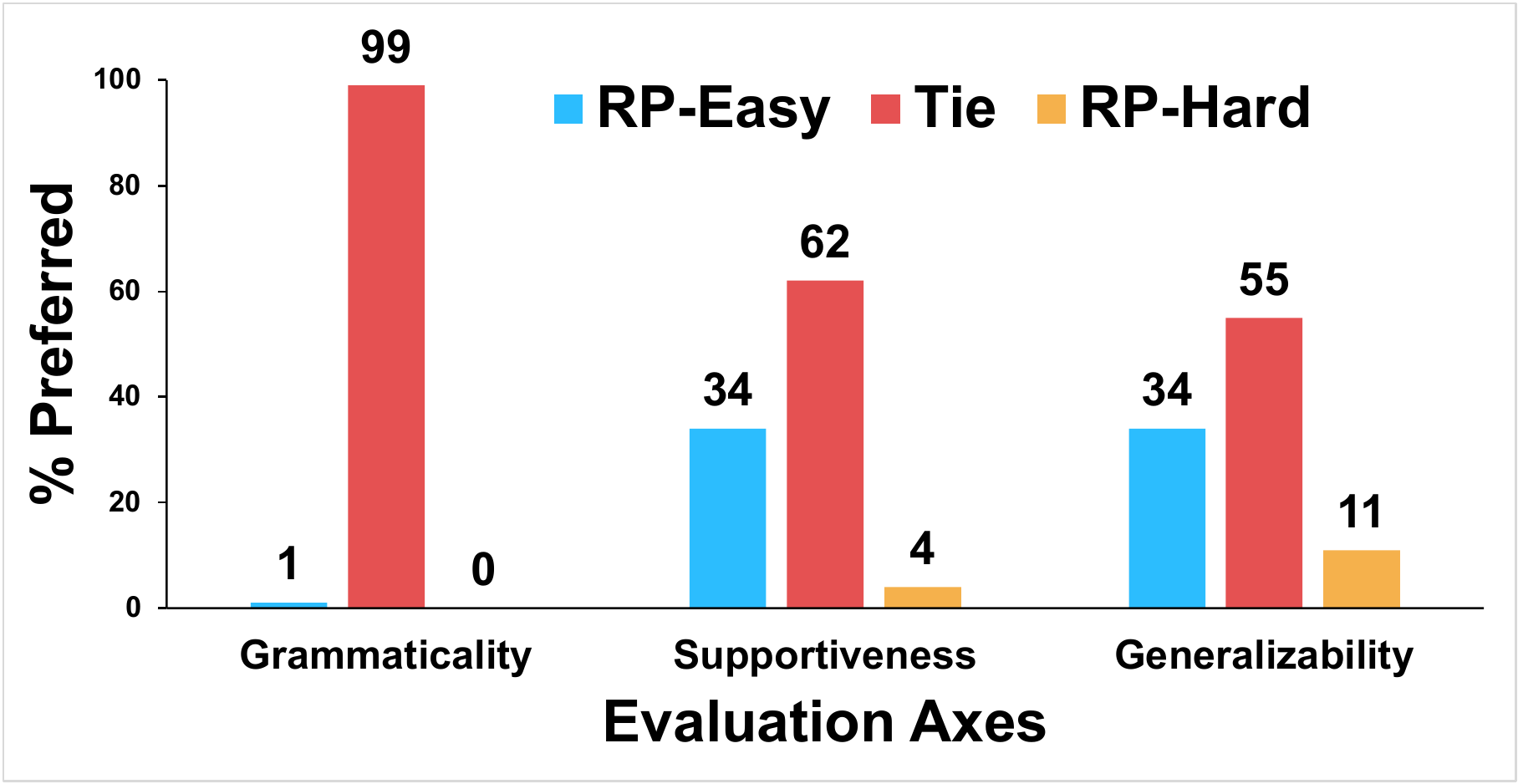}
        \small{(b) Hardness of Retrieval Pool}
    \end{minipage}
    \vspace{-5pt}
    \caption{\label{fig:other_comparisons}Head-to-head comparison of GPT-3 explanations for easy examples by varying the size (400/200) and hardness (easy/hard) of the Retrieval Pool (RP) for choosing in-context examples.}
    \vspace{-15pt}
\end{figure*}

\subsection{RQ2: How much do model explanations vary based on the size and the hardness of the retrieval pool for choosing in-context samples?} 

We investigate RQ2 by conducting two experiments in which we compare the explanations generated by GPT-3 for 100 easy examples. In the first, we vary the size of the retrieval pool (RP) for selecting in-context examples from 400 to 200 while keeping the average hardness constant, and in the second, we vary the hardness of the retrieval pool from easy to hard examples with the size of the pool (400) remaining constant. As shown in Fig.~\ref{fig:other_comparisons}, the grammaticality of the explanations is unaffected in both experiments. However, supportiveness drops when the in-context samples are retrieved from a smaller pool. A larger pool increases the likelihood of having more similar in-context examples to the test sample, and we conclude that similar in-context examples improve the supportiveness of the explanation. We also find that when explaining easy examples, having a retrieval pool of similar easy examples helps the model generate better explanations, possibly because of more similar in-context examples. Combining with RQ1, we conclude that hardness of both in-context and test samples can affect the quality of model explanations.

We also conduct a similar study for comparing the explanation quality of 100 hard test examples by varying the hardness of the retrieval pool. In contrast to easy test examples, we do not observe statistically significant differences in explanation quality for hard examples when the retrieval pool's hardness is varied. In particular, with respect to supportiveness, the win percentages for hard and easy pool are 20\% and 18\% respectively, with remaining 62\% being ties, while for generalizability, they are 33\% and 25\% respectively, with remaining 42\% being ties. We believe that the quality of hard examples may not be sensitive to changes in the in-context examples simply because the corresponding explanations are not very good to begin with.

\subsection{RQ3: Are humans equally good at explaining easy vs. hard examples?}

\begin{figure}
    \centering
        \includegraphics[width=\columnwidth]{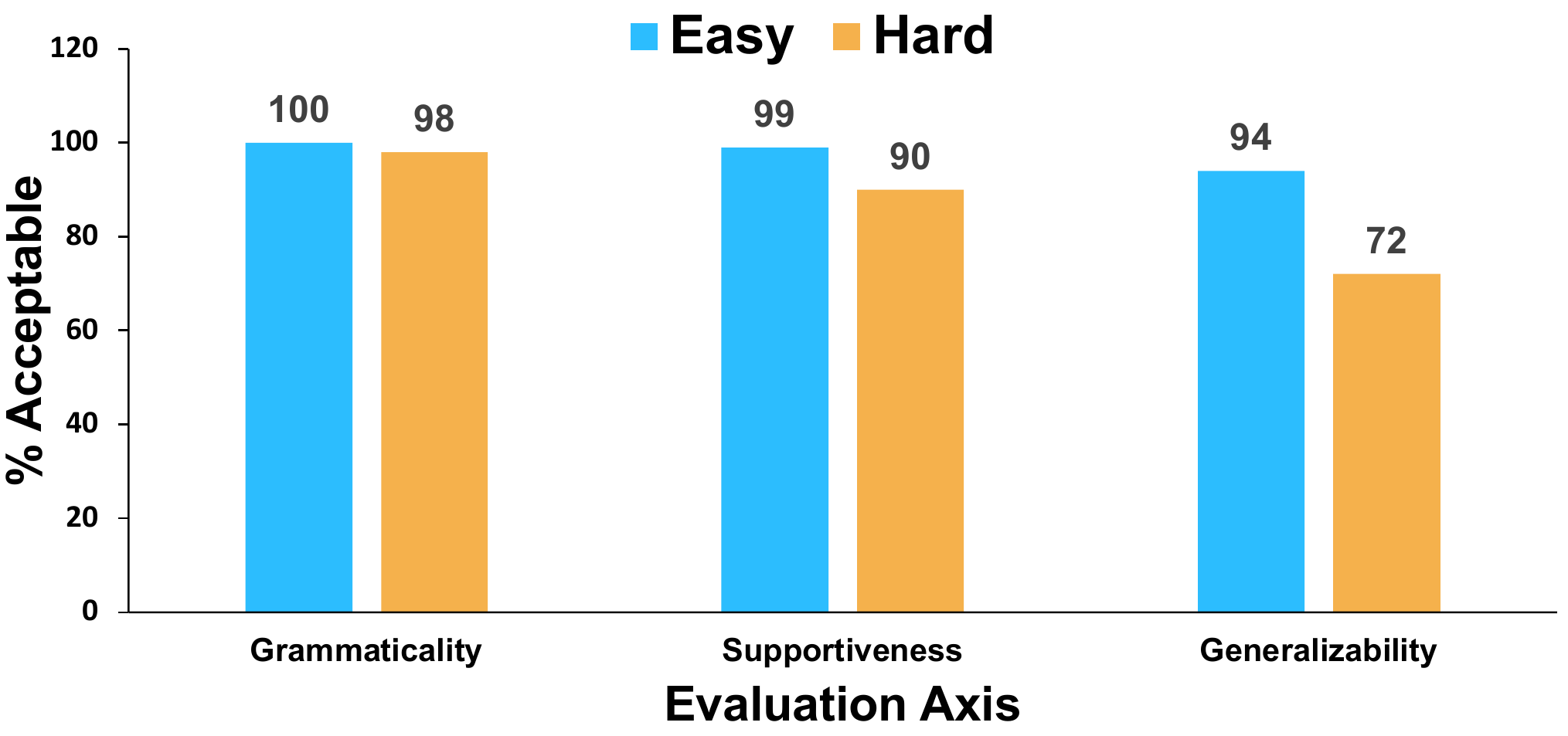} 
        \vspace{-15pt}
    \caption{\label{fig:human_comp}Percentage of acceptable human explanations for easy and hard examples across evaluation axes.}
    \vspace{-5pt}
\end{figure}

In RQ1, we compared the relative performance of model and humans in explaining easy, medium, and hard examples. RQ3 now evaluates the absolute quality of human-written explanations. In particular, we ask the annotators to rate whether the explanations demonstrate \textit{acceptable} grammaticality, supportiveness, and generalizability. Fig.~\ref{fig:human_comp} shows the fraction of \textit{acceptable} human explanations along these three axes for easy and hard examples. We observe that humans also find it hard to write generalizable explanations for some hard examples. Overall, the quality of human explanations is also impacted by the hardness of the samples, although to a lesser extent than GPT-3 since human explanations become clearly preferable to model explanations as hardness increases (RQ1).

\section{Related Work}

There has been significant progress made in recent years on both curating natural language explanation datasets~\cite[\textit{inter alia}]{camburu2018snli, rajani2019explain, brahman2021learning, aggarwal2021explanations} as well as generating them~\cite{rajani2019explain, shwartz2020unsupervised}. Related to the Winograd Schema Challenge, WinoWhy~\cite{zhang2020winowhy} contains explanations only for the WSC273 test set~\cite{levesque2012winograd} and does not follow the structure of our commonsense rule-based explanations, thereby leading to label leakage. Label leakage makes evaluation of explanations harder because supportiveness can become trivial. Our study builds on top of prior works that also generate free-text explanations using in-context learning with GPT-3~\cite{marasovic2021few, wiegreffe2021reframing}. However, our novelty lies in investigating the connection between explainability and sample hardness. A number of concurrent works have also explored free-text explanations for in-context learning in various reasoning tasks~\cite{nye2021show, chowdhery2022palm, wei2022chain, lampinen2022can, wang2022self, ye2022unreliability}, primarily focusing on improving model performance with explanations and not evaluating explanation properties or factors that might influence them.

\section{Conclusion}
We studied the effect of sample hardness on the quality of post-hoc explanations generated by LLMs for data labels. We concluded that while LLM explanations are as fluent as human explanations regardless of the sample hardness, humans are typically better at writing more generalizable explanations and specifically, for hard examples, human explanations are also more supportive of the label. Factors like the hardness and size of the retrieval pool for choosing in-context examples can further impact the explanation quality. We also observe that the generalizability aspect of human explanations drops for harder examples, although by a smaller margin than models.

\section*{Limitations}

The goal of our study is to evaluate how well models explain the data labels and not their own answers for the data points. Hence, both humans and models write or generate post-hoc explanations by conditioning on the gold labels. This also leads us to evaluate the explanations for how acceptable they are to the humans rather than their faithfulness to the model decisions~\cite{wiegreffe2020attention, jacovi2020towards}. The notion of data maps-driven instance difficulty~\cite{swayamdipta2020dataset} is primarily model dependent, and it is conceivable that different choices of models (or model-families) would yield different rank-ordering of data points by hardness. However, we measure the relative hardness of the data points and it is very unlikely that the k-easiest samples for RoBERTa (which is used to estimate sample hardness) will be the k-hardest samples for GPT-3 (which is used to generate explanations) or vice versa. In addition, we find that humans also struggle to explain our estimated `hard' examples. These factors make our results fairly generalizable and future work can explore this direction further. It would also be interesting to see how our results generalize to other forms of explanations in NLP like rationales or structured explanations.

\section*{Acknowledgements}

We thank the reviewers for their helpful feedback and the annotators for their time and effort. This work was supported by NSF-CAREER Award 1846185, NSF-AI Engage Institute DRL-2112635, DARPA MCS Grant N66001-19-2-4031, ONR Grant N00014-18-1-2871, and Google PhD Fellowship. The views contained in this article are those of the authors and not of the funding agency.

\bibliography{custom}
\bibliographystyle{acl_natbib}

\appendix
\section{Crowdsourcing Details}
\label{sec:crowdsource}
\begin{figure}[t]
    \centering
    \includegraphics[width=\columnwidth]{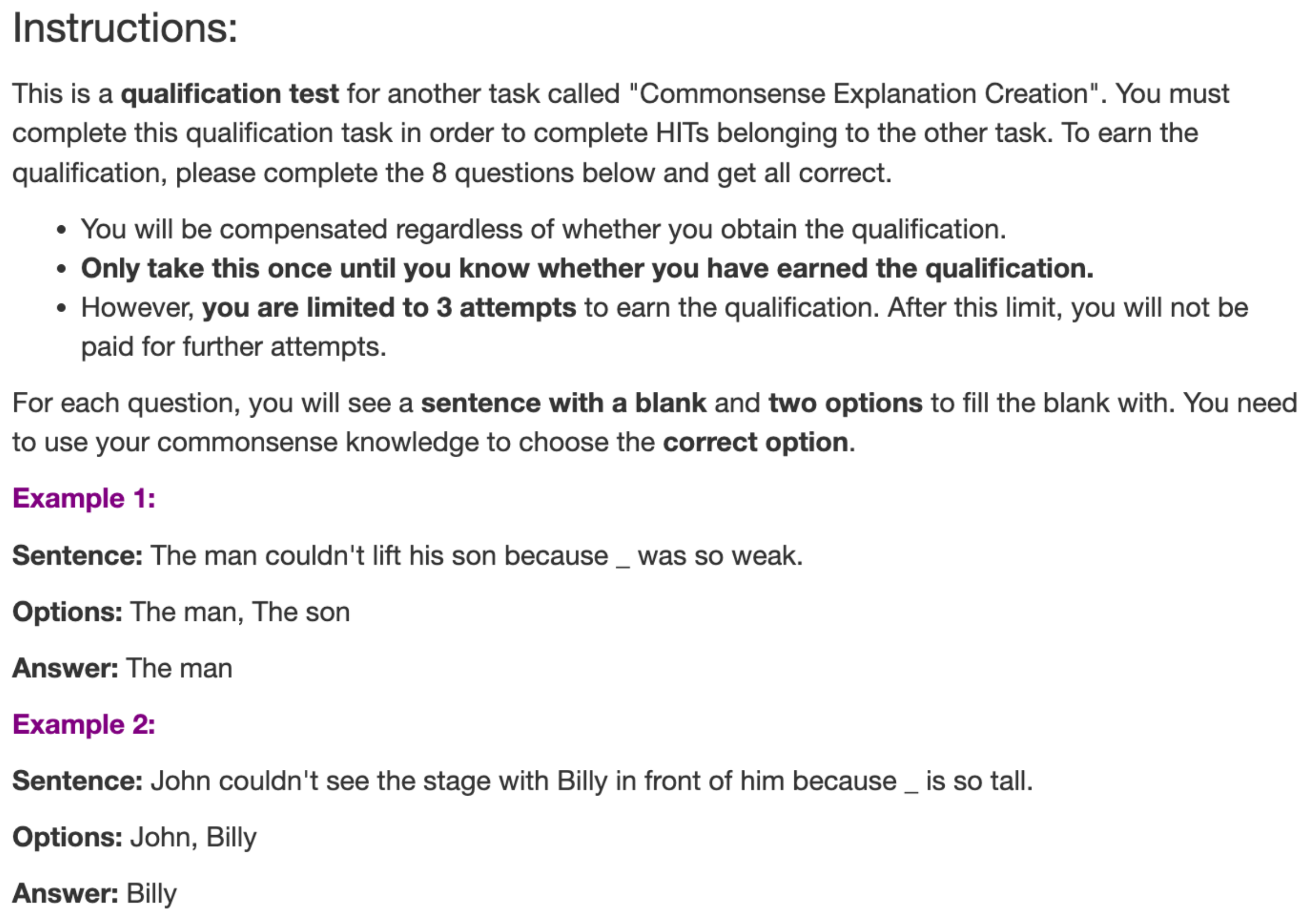}
    \caption{Instructions for the qualification test for writing and verifying explanations for the task of Winograd Schema Challenge.}
    \label{fig:qual}
\end{figure}

\begin{figure}[t]
    \centering
    {\includegraphics[width=\columnwidth]{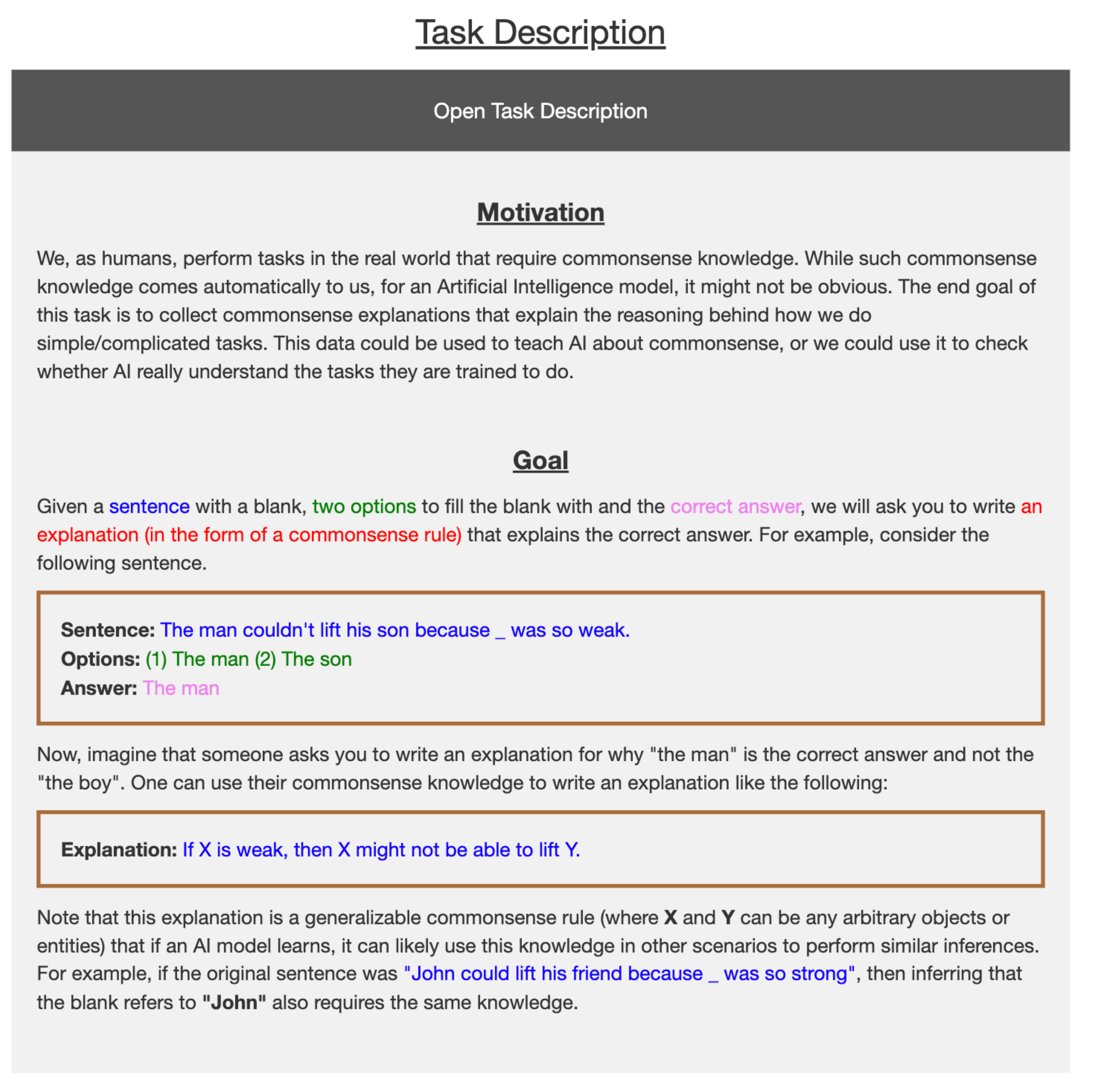}}
    \qquad
    {\includegraphics[width=\columnwidth]{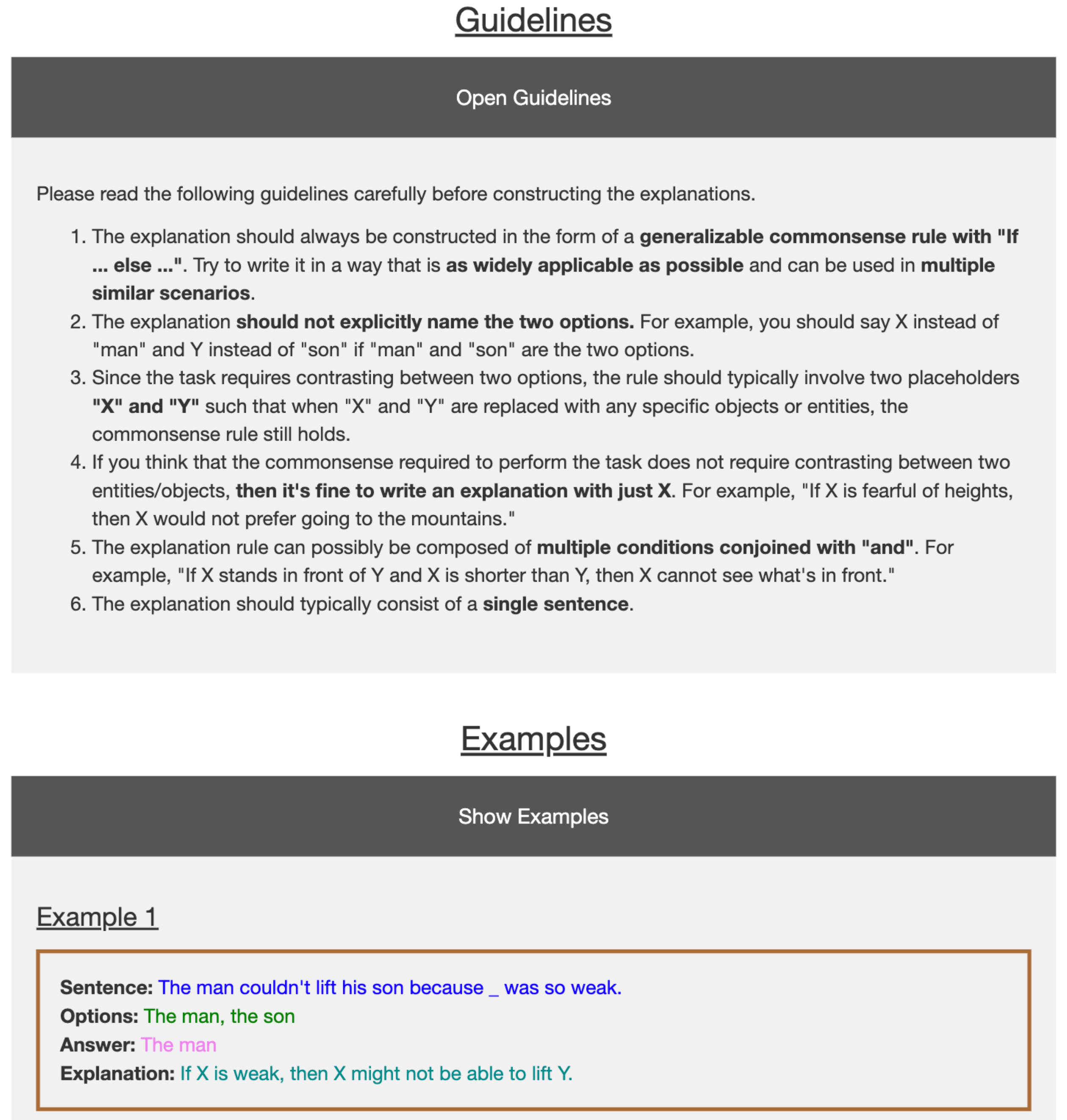}}

    {\includegraphics[width=\columnwidth]{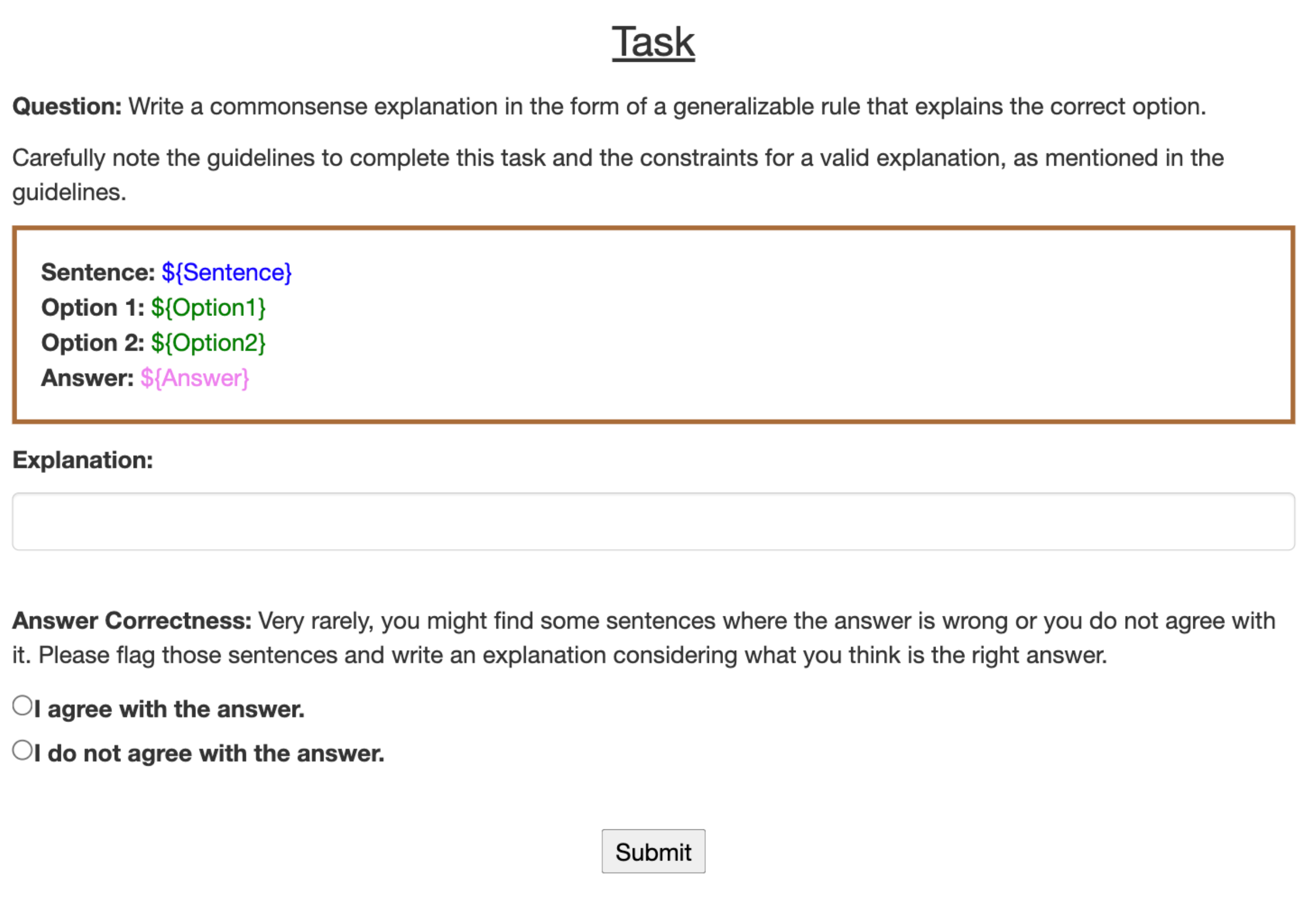}}
    \caption{Explanation Creation Guidelines and Interface on Amazon Mechanical Turk. We ask crowdworkers to follow the guidelines when constructing explanations. We also perform in-browser checks to ensure that the options are not explicitly mentioned in the explanations.}
    \label{fig:explanation_collection}
\end{figure}

\begin{figure}[t]
    \centering
    {\includegraphics[width=\columnwidth]{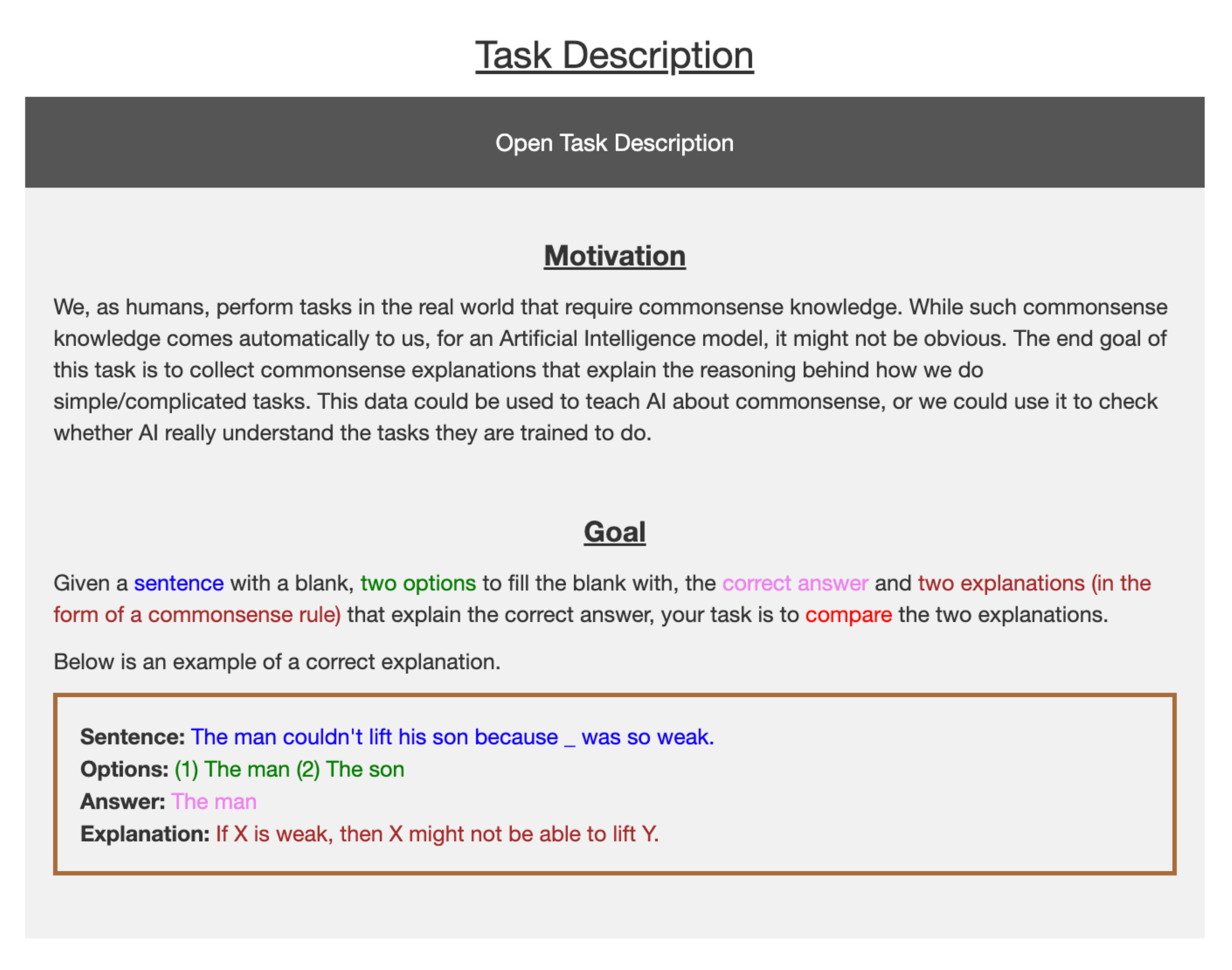}}
    \qquad
    {\includegraphics[width=\columnwidth]{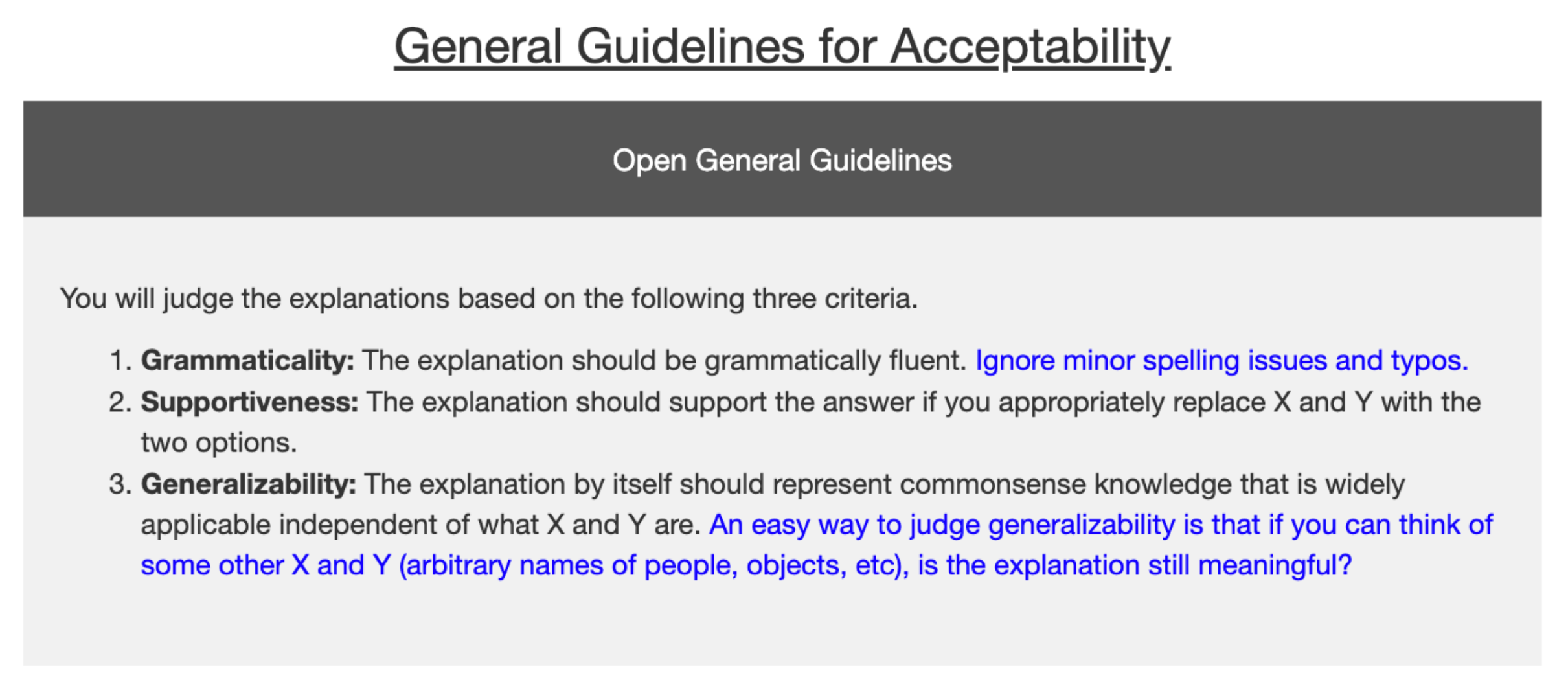}}

    {\includegraphics[width=\columnwidth]{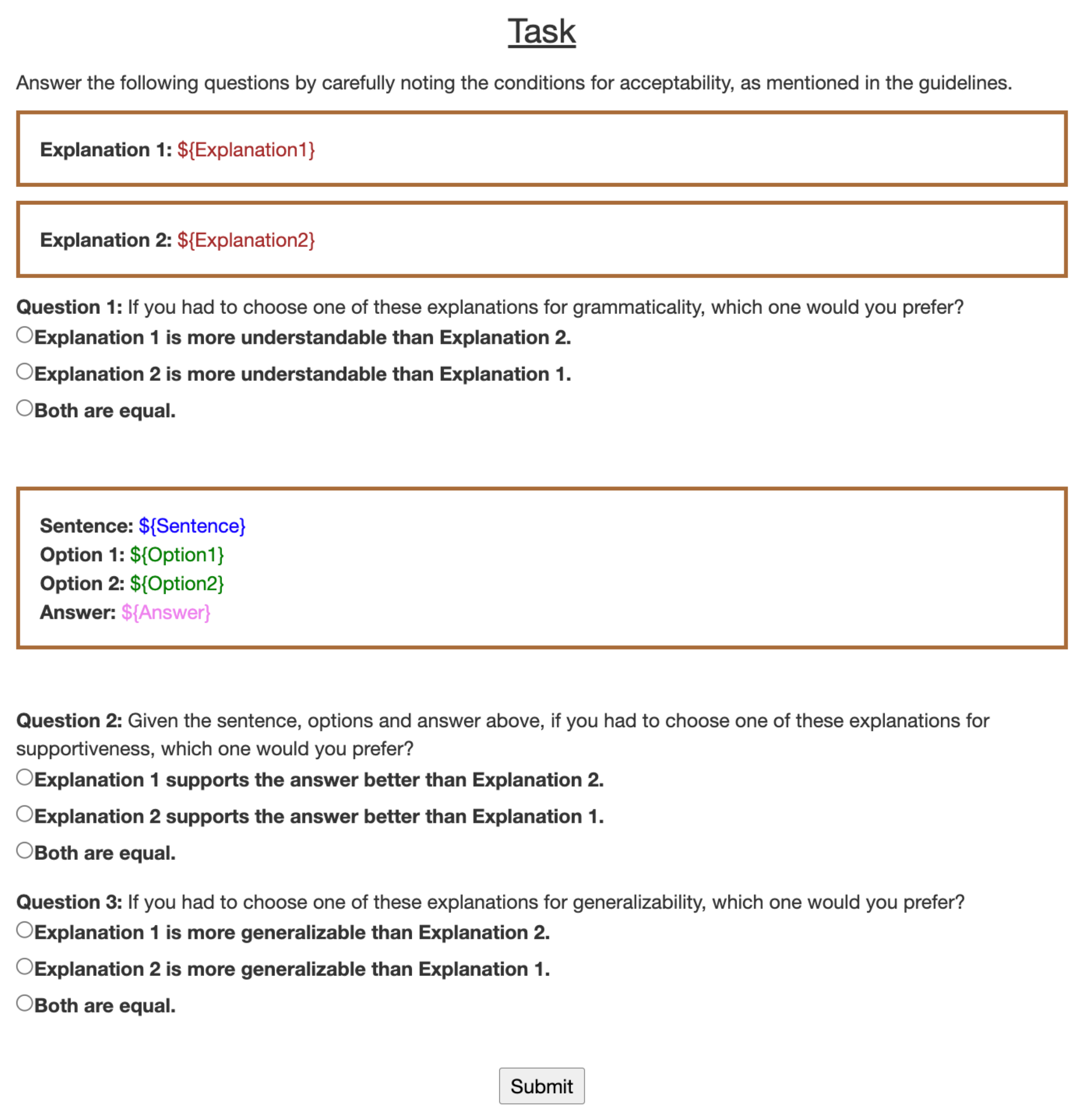}}
    \caption{Explanation Verification Guidelines and Interface on Amazon Mechanical Turk.}
    \label{fig:explanation_verification}
\end{figure}

All our crowdsourcing studies are done on Amazon Mechanical Turk. We select crowdworkers who are located in the US with a HIT approval rate higher than 96\% and at least 1000 HITs approved. We conduct qualification tests before crowdworkers are allowed to write and verify explanations. As shown in Figure~\ref{fig:qual}, it tests the annotator's understanding of the Winograd Schema Challenge by asking to choose the correct option given the sentence and get all questions correct. In Figure~\ref{fig:explanation_collection}, we show the instructions and interface for collecting human-written explanations. Finally, in Figure~\ref{fig:explanation_verification}, we show the interface for explanation verification. We pay annotators \$0.10 for each HIT of explanation construction and \$0.15 for each HIT of explanation verification at an hourly wage of \$12-15. 

\begin{table}[]
    \centering
    \begin{tabular}{ccc}
    \toprule
         & Easy & Hard \\ \midrule
        Grammaticality & 0.63 & 0.61 \\
        Supportiveness & 0.51 & 0.43 \\
        Generalizability & 0.45 & 0.37 \\ \bottomrule
    \end{tabular}
    \caption{Inter-annotator agreement scores (Krippendorff’s $\alpha$~\cite{krippendorff2011computing}) for human evaluation of explanations for easy and hard examples along three evaluation axes.}
    \label{tab:agreement}
\end{table}

\paragraph{Inter-annotator Agreement. }Each explanation is evaluated by three annotators. We report inter-annotator agreement using Krippendorff’s $\alpha$~\cite{krippendorff2011computing}. Despite the subjective nature of our task, we observe moderate agreement scores among annotators, as reported in Table~\ref{tab:agreement}. Perhaps unsuprisingly, we find the agreement score for grammaticality to be the highest and that of generalizability to be the lowest. For supportiveness, we observe an $\alpha$ in the range of 0.4--0.5. Between easy and hard examples, the agreement scores for hard examples are lower, which also shows that these examples are harder for humans to agree on.

\begin{table}[]
\small
\resizebox{\columnwidth}{!}{
\begin{tabular}{l|c}
\toprule
\textbf{Sentence}                                                                                      & \textbf{Options} \\ \midrule
I wanted to buy small tweezer to fit in my wristlet, \\ but they still didn't fit. The \_ were too small. & \centered{tweezer / \underline{wristlet}}   \\ \midrule
The documents contained in the files could not \\ fit properly. The \_ were too large.                    & \centered{\underline{documents} / files}       \\ \midrule
I measured the area in my kitchen, but the stove \\ didn't fit because the \_ was too small.                  & \centered{\underline{kitchen} / stove}    \\ \bottomrule      
\end{tabular}
}
\caption{Examples from the Winogrande dataset requiring the same commonsense knowledge that ``If X is larger than Y, then X does not fit in Y''.
}

\end{table}

\section{Prompting Details}
\label{sec:prompt}
We avoid prompt tuning by largely following~\citet{wiegreffe2021reframing} for prompt construction and choosing a layout that resembles~\citet{wiegreffe2021reframing}'s CommonsenseQA prompt. Following~\citet{liu2022wanli}, we order the in-context samples in increasing order of similarity to the test sample such that the most similar sample is last in the context. All our generated explanations are obtained using the largest ``text-davinci-002'' model of GPT-3\footnote{\url{https://beta.openai.com/docs/models/gpt-3}} with greedy decoding and maximum token limit of $50$. While prior works~\cite{zhao2021calibrate, lu2021fantastically} have shown that in-context learning methods have high variance based on the hyperparameters chosen or the order of examples, we find that our generated explanations are fairly robust to such variations due to their semi-structured form. We also note that finding the most optimal prompt is not the main focus of our work. Instead, we are interested in understanding the connection between explanation quality and sample hardness when other factors like hyperparameters, decoding strategy, etc are kept unaltered.

\begin{figure}[t]
\small
    \centering
    \begin{tabular}{p{\linewidth}}
         \toprule
         Let's explain commonsense questions. \\
         question: \textcolor{blue}{Katrina gave a more dynamic speech than Angela during the presentation since \_ was creative. What does the ``\_'' refer to?} \\
         \textcolor{brown}{Katrina, Angela} \\
         \textcolor{cyan}{Katrina} \\
         why? \textcolor{red}{If X is more creative than Y, X would make the more dynamic speech.} \\
         \#\#\# \\
         question: \textcolor{blue}{Lindsey decided to buy a nice piece of ink wash artwork from Katrina because \_ was a great artist. What does the "\_" refer to?} \\
         \textcolor{brown}{Lindsey, Katrina} \\
         \textcolor{cyan}{Katrina} \\
         why? \textcolor{red}{If X decided to buy an artwork from Y, then Y was a great artist.}
         \#\#\# \\
         question: \textcolor{blue}{Using Rose water as a tan remover worked for Amy but not Katrina because \_ disliked the smell. What does the "\_" refer to?} \\
         \textcolor{brown}{Amy, Katrina} \\
         \textcolor{cyan}{Katrina} \\
         why? \textcolor{red}{If X doesn't like the smell of rose water, then Y is more likely to use the rose water tan remover.}
         \#\#\# \\
         question: \textcolor{blue}{Kyle was not able to have a lavish lifestyle but Lawrence could because \_ had lots of money. What does the "\_" refer to?} \\
         \textcolor{brown}{Kyle, Lawrence} \\
         \textcolor{cyan}{Lawrence} \\
         why? \textcolor{red}{If X has more money than Y, then X would know more about lavish lifestyles.} \\
         \#\#\# \\
         question: \textcolor{blue}{Leslie was able to buy new paint for his house this weekend unlike Nelson, because \_ was wealthy. What does the "\_" refer to?} \\
         \textcolor{brown}{Leslie,	Nelson} \\
         \textcolor{cyan}{Leslie} \\
         why? \textcolor{red}{If X is wealthy, then X will be able to buy new paint, and Y will not.} \\
         \#\#\# \\
         question: \textcolor{blue}{Katrina had the financial means to afford a new car while Monica did not, since \_ had a high paying job.} \\
         \textcolor{brown}{Katrina, Monica} \\
         \textcolor{cyan}{Katrina} \\
         why? \\
         \bottomrule
    \end{tabular}
    \caption{Example of a prompt with five in-context samples for Winogrande. Each sample consists of the \textcolor{blue}{question}, \textcolor{brown}{two options}, \textcolor{cyan}{the correct answer} and \textcolor{red}{an explanation} in the form of a generalized commonsense rule. The in-context samples are arranged in increasing order of similarity to the test sample. GPT-3 generates a free-text explanation for the current sample:  \textit{If X has a higher paying job than Y, then X would have more money to afford a new car.}}
    \label{tab:prompt_example}
\end{figure}

\begin{table*}[t]
\centering
\small
\resizebox{\textwidth}{!}{
\begin{tabular}{p{0.35\linewidth} | p{0.35\linewidth} | p{0.3\linewidth} }
\toprule
\textbf{Sample} & \textbf{1st Similar Example} & \textbf{2nd Similar Example} \\ \midrule
Katrina had the financial means to afford a new car while Monica did not, since \_ had a high paying job. &	Leslie was able to buy new paint for his house this weekend unlike Nelson, because \_ was wealthy. & Kyle was not able to have a lavish lifestyle but Lawrence could because \_ had lots of money. \\ \midrule
Bill's new houseboat he purchased would not fit in his garage, the \_ was too small. & I tried to set the plant in the pot, but it didn't work because the \_ was too large. &	The bottles supplied is not enough to collect the water. The \_ is too much. \\ \midrule
She had a cold and decided to ditch the vitamins and use medicines, because the \_ were less effective. & My cousin preferred the treatments over the procedures because the \_ were better for your health. &	I removed beef from my diet and added pork, as the \_ turned out to be unhealthy. \\
\bottomrule      
\end{tabular}
}
\caption{Examples of similar samples retrieved for in-context learning. The similar examples require similar commonsense reasoning for inferring the correct answer.\label{tab:top_k_similar}}

\end{table*}

\section{Examples of Similar Examples Retrieved for In-Context Learning}
\label{sec:top_k_samples}
We find that our simple method of using sentence embeddings to retrieve top-k similar examples for in-context learning works well in practice. In Table~\ref{tab:top_k_similar}, we show some representative examples, demonstrating the presence of similar commonsense knowledge between the test sample and the top-2 similar samples.

\begin{table*}[t]
\centering
\resizebox{\textwidth}{!}{
\begin{tabular}{p{0.4\linewidth} | l | l | p{0.3\linewidth} | p{0.3\linewidth} }
\toprule
\textbf{Sentence} & \textbf{Options} & \textbf{Answer} & \textbf{GPT-3 Explanation} &\textbf{Reason} \\ \midrule
Matthew is trying to make William's wiener dog happy, because \_ is left alone with it. &	Matthew/William &	Matthew &  If X is left with Y then X is responsible for Y. & Unable to understand what X and Y are in the context. \\ \midrule
The couple couldn't decide whether to get a new refrigerator or new laundry machine. They went with the former because their old \_ was dilapidated. &	refrigerator/laundry machine &	refrigerator &  If X is old and Y is new, you will go with X. & Doesn't explain the core concept that old things must be replaced first. \\ \midrule
Jean preferred to use the scrub over the body soap to clean off dead skin because the \_ was harsher. &	scrub/body soap &	scrub &	 If X is harsher than Y, then X is preferred. & Misses the important part of ``preferred for cleaning'' without which the explanation is incomplete and not generalizable.\\ \midrule
We tried to get the bookcase through the door, but the \_ was not narrow enough. &	door/bookcase &	bookcase &  If X is not narrow enough, then Y can't get through. & Wrong explanation as it does not support the label. \\ \midrule
Because Lindsey wanted to make a mess and Victoria did not, \_ cracked her eggs and opened them over the bowl. &	Lindsey/Victoria &	Victoria &	 If X wanted to make a mess, and Y didn't, then Y would do the opposite of X. & Explanation not relevant to the context. \\
\bottomrule      
\end{tabular}
}
\caption{A few representative examples of bad explanations generated by GPT-3 for hard examples.\label{tab:errors_hard}}

\end{table*}

\section{Analysis of GPT-3-generated Explanations for Hard Samples}
\label{sec:analysis-gpt3}
In Table~\ref{tab:errors_hard}, we show more examples of bad explanations generated by GPT-3 for some of the hard examples. While the model is able to learn the semi-structured nature of the explanations, it often makes mistakes in identifying what X and Y are (first example), misses the core reasoning concept (second and third examples) or are non-contextual (last example), thereby either not properly supporting the label or completing refuting the label (fourth example). Consequently, the `generalizability' aspect of these explanations also suffer.

\end{document}